\documentclass[runningheads]{llncs}
\usepackage[T1]{fontenc}
\usepackage{graphicx}
\usepackage{blindtext}
\usepackage{hyperref}

\usepackage[utf8]{inputenc}
\usepackage{url}
\usepackage{amsmath}
\usepackage{booktabs}
\usepackage{algorithm}
\usepackage{algorithmic}

\newcommand{\orcidAuthorOne}	{\href{https://orcid.org/0009-0006-3215-3775}{\protect\includegraphics[scale=0.045]{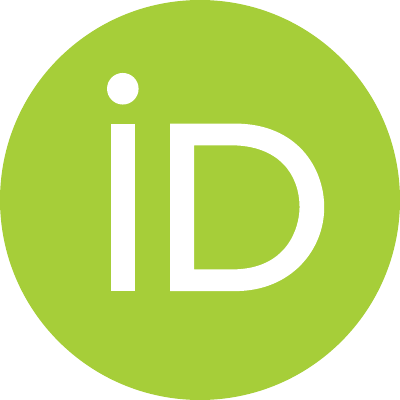}}}
\newcommand{\orcidAuthorTwo}	{\href{https://orcid.org/0009-0001-2156-6543}{\protect\includegraphics[scale=0.045]{img/orcid}}}
\newcommand{\orcidAuthorThree}	{\href{https://orcid.org/0000-0003-4181-9285}{\protect\includegraphics[scale=0.045]{img/orcid}}}
\newcommand{\orcidAuthorFour}	{\href{https://orcid.org/0000-0002-6838-6320}{\protect\includegraphics[scale=0.045]{img/orcid}}}
\newcommand{\orcidAuthorFive}	{\href{https://orcid.org/0009-0008-1070-6759}{\protect\includegraphics[scale=0.045]{img/orcid}}}
\newcommand{\orcidAuthorSix}	{\href{https://orcid.org/0009-0002-4610-2662}{\protect\includegraphics[scale=0.045]{img/orcid}}}
\newcommand{\orcidAuthorSeven}	{\href{https://orcid.org/0009-0002-6572-6892}{\protect\includegraphics[scale=0.045]{img/orcid}}}
\newcommand{\orcidAuthorEight}	{\href{https://orcid.org/0000-0002-6866-0799}{\protect\includegraphics[scale=0.045]{img/orcid}}}
\newcommand{\orcidAuthorNine}	{\href{https://orcid.org/0009-0008-6067-2702}{\protect\includegraphics[scale=0.045]{img/orcid}}}
\newcommand{\orcidAuthorTen}	{\href{https://orcid.org/0000-0002-7533-2347}{\protect\includegraphics[scale=0.045]{img/orcid}}}
\newcommand{\orcidAuthorEleven}	{\href{https://orcid.org/0000-0001-6257-7399}{\protect\includegraphics[scale=0.045]{img/orcid}}}

\begin{document}
\title{TextBFGS: A Case-Based Reasoning Approach to Code Optimization via Error-Operator Retrieval}
\titlerunning{TextBFGS}

\author{
Zizheng Zhang\inst{1}\orcidAuthorOne \and
Yuyang Liao\inst{1}\orcidAuthorTwo \and
Chen Chen\inst{3}\orcidAuthorThree \and
Jian He\inst{1}\orcidAuthorFour \and
Dun Wu\inst{1}\orcidAuthorFive \and
Qianjin Yu\inst{1}\orcidAuthorSix \and
Yanqin Gao\inst{1}\orcidAuthorSeven    \and
Jin Yang\inst{2}\orcidAuthorEight \and
Kailai Zhang\inst{2}\orcidAuthorNine \and \\
Eng Siong Chng\inst{3}\orcidAuthorTen \and
Xionghu Zhong\inst{4}\orcidAuthorEleven
}
\authorrunning{Z. Zhang et al.}

\institute{
Zhongxing Telecom Equipment (ZTE), China \\
\email{zzz128@alumni.pku.edu.cn} \\
\url{https://github.com/TzuchengChang} \and
China Mobile, China \and
Nanyang Technological University (NTU), Singapore \and
Hunan University, China
}

\maketitle

\begin{abstract}
Iterative code generation with Large Language Models (LLMs) can be viewed as an optimization process guided by textual feedback. However, existing LLM self-correction methods predominantly operate in a stateless, trial-and-error manner akin to first-order search, failing to leverage past problem-solving experiences. To bridge this gap, we introduce TextBFGS, a Case-Based Reasoning (CBR) framework inspired by the Quasi-Newton optimization method. Instead of retrieving raw, unstructured textual instances, TextBFGS maintains a dynamic Case Base of historical "Error-to-Operator" correction trajectories to approximate the semantic curvature (inverse Hessian matrix) of the task. Specifically, given a textual error feedback (the target problem), TextBFGS retrieves analogous historical correction patterns (Retrieve) and applies these abstract operators to refine the current code (Reuse/Revise). Furthermore, successful adaptations are continuously retained back into the Case Base (Retain), enabling a self-evolving system. Empirical evaluations on Python code optimization tasks (HumanEval, MBPP) demonstrate that TextBFGS significantly outperforms stateless baselines. It achieves superior pass rates with fewer model calls, establishing an efficient, experience-driven paradigm for LLM-based code optimization.

\keywords{Case-Based Reasoning \and Large Language Models \and Code Optimization \and Retrieval Augmented Generation \and Experience Reuse}
\end{abstract}

\section{Introduction}

Large Language Models (LLMs) have demonstrated remarkable capabilities, particularly in code generation and automated optimization~\cite{sahoo2024systematic,ma2025agentic}. However, their performance heavily relies on iterative refinement rather than a single forward pass. Consequently, code optimization has evolved from a manual trial-and-error process into a systematic optimization problem. Early works such as OPRO~\cite{yang2023large} and MIPRO~\cite{opsahl2024optimizing} explored automatic prompt generation using LLMs as meta-optimizers. While TextGrad~\cite{yuksekgonul2024textgrad}, have formalized this paradigm by treating LLM-generated textual error feedback as ``gradients'' ($\nabla$), enabling the optimization of discrete code variables through iterative backpropagation.

Despite these advancements, existing textual optimizers still predominantly operate as first-order methods, conceptually equivalent to Stochastic Gradient Descent (SGD)~\cite{amari1993backpropagation}. While effective in simple scenarios, first-order approaches in the discrete semantic space suffer from significant limitations. First, they are stateless: each optimization run begins tabula rasa, failing to leverage past problem-solving experiences (i.e., historical cases). Second, they neglect the semantic curvature of the optimization landscape. In numerical optimization, the Hessian matrix describes local geometry, indicating how aggressively parameters should be updated. In code optimization, naive gradient descent treats all errors with uniform step logic, often leading to inefficient ``zig-zagging''---where the model operates without addressing the underlying misconceptions---or requiring excessive token consumption to converge.

To address the statelessness of SGD, recent frameworks like REMO~\cite{wu2025reflection} have attempted to incorporate case-based memory modules by retrieving successful correction traces. However, these methods typically rely on input similarity for retrieval (e.g., retrieving similar code problems). We argue that this surface-level retrieval limits generalization. A ``boundary error'' in a sorting algorithm shares the same structural defect as one in a pathfinding algorithm, even if their problem descriptions are textually disjoint. By binding retrieval to surface-level content rather than the error dynamics, existing memory-augmented methods fail to facilitate effective transfer learning across different task domains.

In this paper, we introduce TextBFGS, a Case-Based Reasoning (CBR) framework for code optimization, conceptually inspired by the Quasi-Newton optimization method, which approximates the inverse Hessian matrix using historical gradient differences. TextBFGS approximates semantic curvature via a novel Gradient-Operator retrieval mechanism. Instead of retrieving input based on problem similarity, we construct a dynamic Case Base mapping specific feedback patterns (gradients) to abstract modification strategies (operators). Crucially, TextBFGS operates in a One-Pass manner. Upon receiving feedback (the target problem), it retrieves high-order correction operators from its trajectory memory (Retrieve) and fuses the gradient generation (perception) and variable modification (update) into a single inference step (Reuse/Revise). This design not only injects second-order guidance to accelerate convergence but also significantly reduces the computational overhead compared to multi-step pipelines.

Our contributions are summarized as follows:
\begin{itemize}
    \item \textbf{Gradient-Operator
Retrieval:} We bridge the gap between numerical second-order optimization and Case-Based automated code optimization, proposing TextBFGS and a robust baseline TextBFGS-REMO, adapting the existing input-based retrieval method into our efficient One-Pass architecture to enable a rigorous comparison between different retrieval paradigms.
    \item \textbf{One-Pass Update \& Case Retention:} We propose the One-Pass update mechanism, which unifies feedback diagnosis and operator application into a single LLM inference step, significantly reducing computational overhead. This is coupled with a Case Retention loop where successful optimization trajectories, consisting of gradients and abstract operators, are dynamically injected back into the memory. This allows the optimizer to continuously refine its Case Base. 
    \item \textbf{Empirical Superiority \& Efficiency:} Experiments on rigorous code optimization benchmarks demonstrate the superiority and efficiency of TextBFGS. It achieves superior pass rates (+20.5\% in MBPP) with fewer token counts (-50.6\% in HumanEval) compared to TextGrad-based approaches and exhibits strong cross-domain transferability (+16.2\% in MBPP) compared to input-based retrieval, effectively applying debugging logic learned from one Case Base to another Case Base.
\end{itemize}

\section{Related Work}
\subsection{Zero-Order Approaches: Stateless Heuristic Search}

Early approaches treated LLMs as black-box functions, navigating the optimization landscape through iterative mutation and selection without explicit directional guidance (gradients). OPRO~\cite{yang2023large} pioneered using LLMs as meta-optimizers to generate prompts based on the scalar trajectory of past scores. DSPy~\cite{khattab2024dspy} introduced a declarative programming model where the MIPRO~\cite{opsahl2024optimizing} optimizer employs Bayesian-inspired search strategies. These methods, including PromptBreeder~\cite{fernando2023promptbreeder} and EvoPrompt~\cite{guo2023connecting}, rely fundamentally on blind trial-and-error. They suffer from high sample complexity because they lack a diagnostic mechanism to explain why a candidate failed, relying solely on scalar rewards to guide the search, entirely lacking the ability to reuse past debugging experiences.

\subsection{First-Order Approaches: Feedback-Driven Debugging}

To introduce directionality, recent works~\cite{madaan2023self} formalize the use of natural language feedback as Textual Gradients~\cite{yuksekgonul2024textgrad}, treating the LLM's critique of an output as a gradient ($\nabla$), conceptually updating the code variable akin to SGD~\cite{amari1993backpropagation}.

Although originally framed as evolutionary, AlphaEvolve~\cite{novikov2025alphaevolve} and GEPA~\cite{agrawal2025gepa} also fall under this feedback-driven paradigm. Unlike blind mutation, they utilize feedback to guide its genetic crossover and mutation operations, effectively performing population-based gradient descent. While these methods introduce directionality, they remain strictly first-order and stateless.

Besides, TextGrad's ``momentum'' variant is a simple linear aggregation of past critiques, lacking the geometric interpretation of curvature found in code optimization. Consequently, first-order methods treat the semantic landscape as Euclidean and flat, often leading to inefficient ``zig-zagging'' in complex code debugging tasks where the descent direction does not point to the global optimum.

\subsection{Memory-Augmented Optimization: Towards Case-Based Reasoning}

Addressing the statelessness of gradient descent, recent research integrates memory (conceptually akin to a Case Base in CBR) to approximate higher-order geometric information. In numerical optimization, the Hessian matrix (curvature) captures the relationship between parameter updates ($\Delta x$) and gradient changes ($\Delta g$). In the semantic space, historical optimization trajectories---tuples of mistakes and their successful corrections---serve as cases (problem-solution pairs) that proxy this curvature information.

Reflexion~\cite{shinn2023reflexion}, REVOLVE~\cite{zhang2024revolve}, and HessianGrad~\cite{zhanghessiangrad} utilize short-term memory or meta-prompt optimization~\cite{suzgun2024meta} to avoid repeating immediate errors. However, it limits the fidelity of Hessian approximation. If the optimization stagnates early or fails to discover valid paths, the history remains sparse. Consequently, the optimizer cannot derive curvature information from an empty buffer, leaving it blind to the landscape's geometry.

To overcome this limitation, REMO~\cite{wu2025reflection} couples TextGrad with a Retrieval-Augmented Generation (RAG)~\cite{lewis2020retrieval} module acting as a preliminary Case Base. By retrieving past corrections, REMO implicitly approximates the inverse Hessian, guiding the optimizer with historical structural wisdom. However, REMO relies on \textit{input semantic similarity} for retrieval, restricting generalization across domains. Our TextBFGS bridges this gap by shifting the retrieval paradigm from surface-level inputs to \textit{error dynamics (gradients)}, enabling experience-driven code optimization with robust cross-task transferability.

\section{Methodology}

\begin{figure}[tb]
\centering
\includegraphics[width=\textwidth]{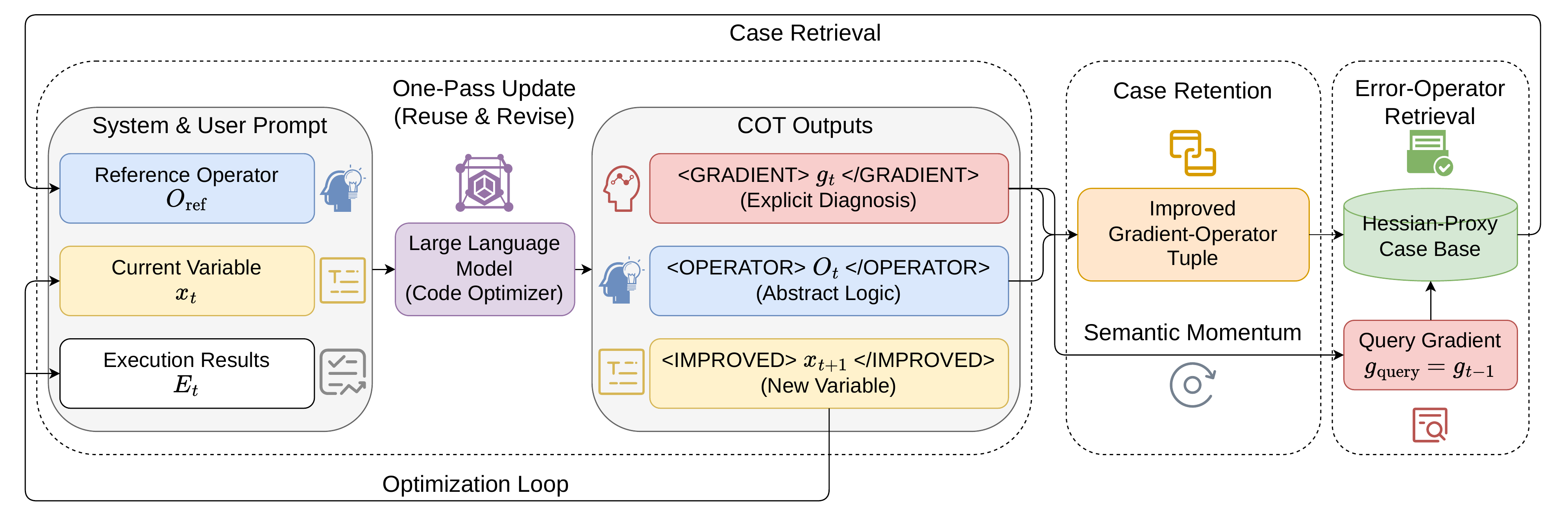}
\caption{The schematic overview of the proposed TextBFGS framework, explicitly mapped to the Case-Based Reasoning (CBR) cycle. (1) \textbf{Case Retrieval (Retrieve):} Instead of retrieving based on input similarity, TextBFGS retrieves abstract optimization operators $\mathcal{O}_{\text{ref}}$ based on error/gradient similarity from the Case Base. (2) \textbf{One-Pass Update (Reuse \& Revise):} The LLM receives the current variable $x_t$, execution results $E_t$, and retrieved operators $\mathcal{O}_{\text{ref}}$ to simultaneously generate an explicit diagnosis (\texttt{<GRADIENT>}), a general correction rule (\texttt{<OPERATOR>}), and the improved variable (\texttt{<IMPROVED>}) in a single inference step. (3) \textbf{Case Retention (Retain):} Upon successful validation (Evaluation), the new error-operator case is injected back into the Case Base, allowing the system to self-evolve and accumulate debugging wisdom.}
\label{fig:onepass}
\end{figure}

In this section, we formalize the problem of code optimization and introduce TextBFGS as a Case-Based Reasoning (CBR) system. We first briefly review the stateless first-order search to highlight its limitations in the semantic space, and then derive our Retrieval-Approximated Quasi-Newton update rule, explicitly mapping it to the classic CBR cycle, as shown in Figure\ref{fig:onepass} and Algorithm\ref{alg:textbfgs}.

\subsection{Problem Formulation: Error Feedback as Target Problem}

We consider the optimization of a discrete executable variable $x$ (e.g., a code snippet) to maximize an objective function $f(x)$, which represents the performance score on a downstream task (e.g., unittest passrate). Since $f(x)$ is non-differentiable with respect to discrete tokens, we adopt the Textual Gradient~\cite{yuksekgonul2024textgrad}.

Let $g_t$ be the error feedback signal generated by an LLM-based evaluator, which serves as the negative gradient $\nabla f(x_t)$. In the context of CBR, $g_t$ functions as the \textit{Target Problem Description} that needs to be resolved. In standard TextGrad, the update follows a stateless SGD scheme:

\begin{equation}
x_{t+1} = \text{LLM}_{\text{update}}(x_t, g_t)
\end{equation}

where $\text{LLM}_{\text{update}}$ is an LLM call that applies the feedback to the variable without consulting any historical memory.

\begin{figure}[tb]
\centering
\includegraphics[width=0.6\columnwidth]{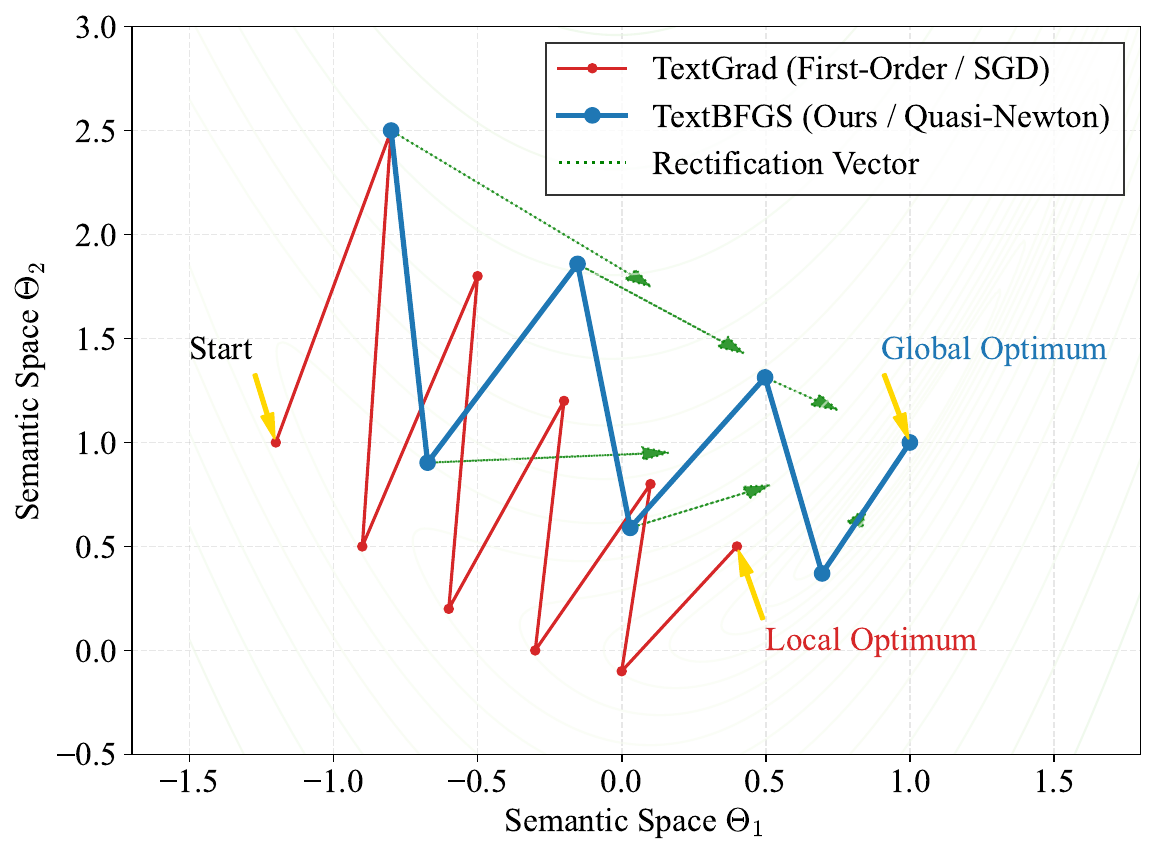}
\caption{Schematic visualization of debugging trajectories in a semantic space. TextGrad (Red) operates as a stateless first-order optimizer, exhibiting inefficient oscillation and getting trapped in a local optimum. In contrast, TextBFGS (Blue) adopts a Quasi-Newton inspired CBR approach. It utilizes prior error dynamics to query the Hessian-Proxy Case Base for historical adaptation rules. This case reuse acts as a structural rectification mechanism (Green), allowing TextBFGS to bypass superficial edits, dampen oscillations, and converge faster to a robust solution.}
\label{fig:curvature}
\end{figure}

\begin{algorithm}[tb]
\caption{TextBFGS Case-Based Optimization Loop}
\label{alg:textbfgs}
\textbf{Input:} Initial variable $x_0$, Objective metric $f(\cdot)$, Hessian-Proxy Case Base $\mathcal{M}$\\
\textbf{Output:} Optimized variable $x_{\text{best}}$
\begin{algorithmic}[1]
\FOR{step $t = 0, 1, \dots, T$}
    \IF{$t > 0$}
    \STATE Query Error/Gradient: $g_t^{\text{query}} \leftarrow g_{t-1}$
    \ENDIF
    \STATE Execute: Run $f(x_{t})$ to get execution results $E_t$.
    \STATE \textit{\# CBR Phase 1: Retrieve}
    \STATE Case Retrieval:
    \STATE \quad $\mathcal{O}_{\text{ref}} \leftarrow \text{Retrieve}(\mathcal{M}, \text{query}=g_t^{\text{query}}, \text{num}=k)$
    \STATE \textit{\# CBR Phase 2 \& 3: Reuse and Revise}
    \STATE One-Pass Update:
    \STATE \quad $\text{Prompt} \leftarrow \mathcal{P}(x_t, E_t, \mathcal{O}_{\text{ref}})$
    \STATE \quad $g_t, \mathcal{O}_{t}, x_{t+1} \leftarrow \text{LLM}(\text{Prompt})$ 
    \IF{$f(x_{t+1})$ is better than $f(x_{t})$}
    \STATE \textit{\# CBR Phase 4: Retain}
    \STATE Case Retention:
    \STATE \quad $\mathcal{M} \leftarrow \mathcal{M} \cup \{(g_t, \mathcal{O}_{t})\}$
    \ENDIF
\ENDFOR
\RETURN Optimized variable $x_{\text{best}}$
\end{algorithmic}
\end{algorithm}

\subsection{Semantic Curvature and the Newton Step as Case Adaptation}

A critical limitation of the SGD update above is its assumption of an isotropic optimization landscape. In reality, the semantic space of code debugging exhibits high curvature~\cite{ethayarajh2019contextual}. First-order methods, lacking knowledge of this geometry and past experiences, often overshoot or oscillate, as shown in Figure\ref{fig:curvature}.

In numerical optimization, Quasi-Newton methods such as BFGS~\cite{liu1989limited} approximate the inverse Hessian matrix to capture local curvature information, enabling faster convergence. Drawing inspiration from BFGS, we conceptually adapt this philosophy to design our CBR system for discrete semantic domain.

To accelerate convergence, we aim to perform a Quasi-Newton step:

\begin{equation}
x_{t+1} = \text{LLM}_{\text{update}}(x_t, H^{-1}_t, g_t)
\end{equation}

where $H^{-1}_t$ represents the inverse Hessian matrix. In numerical optimization, $H^{-1}_t$ acts as a geometric transformation matrix. In our discrete code optimization, explicitly calculating this matrix is mathematically intractable. Instead, we propose that a \textit{Case Base} of historical optimization operators acts as a conceptual approximation of the inverse Hessian. 

We justify this conceptual approximation through functional equivalence. Just as numerical BFGS accumulates past gradient differences and step updates to estimate curvature, our Case Base accumulates historical errors ($g$) and successful abstract operators ($\mathcal{O}$). Retrieving an operator for a specific error applies a structural transformation to the code edit, analogous to how $H^{-1}$ rotates a gradient to correct for landscape geometry. Furthermore, if a current problem exactly matches a historical case, reusing its operator perfectly reconstructs the successful update, effectively satisfying a discrete analog of the Quasi-Newton secant condition ($H \Delta g = \Delta x$).

Since the current gradient $g_t$ is unknown prior to generation, we utilize the ``Semantic Momentum'' derived from the previous step ($g_{t-1}$) to estimate this curvature matrix, assuming that error directions exhibit local persistence~\cite{ethayarajh2019contextual}:

\begin{equation}
\label{eq:hessian_approx}
H^{-1}_t \approx \mathcal{T}(g_{t-1}; \mathcal{M})
\end{equation}

Here, $\mathcal{T}$ denotes the retrieval-based transformation function and $\mathcal{M}$ denotes the Hessian-Proxy Case Base. Unlike standard RAG which retrieves raw text, our $\mathcal{T}$ retrieves a \textit{successful adaptation rule} (operator) based on historical error dynamics. Consequently, the inference of TextBFGS can be formalized as:

\begin{equation}
\label{eq:update_step}
\Delta x_t \approx \underbrace{\mathcal{T}(g_{t-1})}_{\text{Hessian Proxy / Case Retrieval}} \otimes \underbrace{g_t}_{\text{Current Target Problem}}
\end{equation}

where $\otimes$ represents the LLM's internal reasoning process that applies the retrieved abstract operator ($\mathcal{T}$) to the specific current diagnosis ($g_t$) to generate the final variable update.

\subsection{Approximating Hessian via Error-Operator Retrieval}

TextBFGS approximates $H^{-1}_t$ using a Case-Based retrieval mechanism~\cite{gao2025devil}. We construct a Hessian-Proxy Case Base (HPCB), denoted as $\mathcal{M}$, which stores pre-learned successful optimization trajectories as cases (tuples):

\begin{equation}
\mathcal{M} = \{(g_i, \mathcal{O}_i)\}_{i=1}^{N}
\end{equation}

Here, $g_i$ is the textual gradient (the historical error description / Problem), and $\mathcal{O}_i$ is the Optimization Operator---the abstract logic used to resolve $g_i$ (the adaptation rule / Solution), rather than the raw code difference.

\textbf{The Retrieval Process:}
We utilize the previous gradient $g_{t-1}$ as the current query $g_t^{\text{query}}$ based on Semantic Momentum, facilitating One-Pass inference. We then query $\mathcal{M}$ to find the $k$ nearest neighbors~\cite{zhang2016introduction} based on error similarity:

\begin{equation}
\begin{split}
\mathcal{O}_{\text{ref}} &= \text{Retrieve}(\mathcal{M}, g_t^{\text{query}}, k) \\
&= \operatorname*{Top-}k_{(g_i, \mathcal{O}_i) \in \mathcal{M}} \left( \text{CS}(\text{Embed}(g_t^{\text{query}}), \text{Embed}(g_i)) \right)
\end{split}
\end{equation}

where $\text{Embed}(\cdot)$ is a vector embedding, and $\text{CS}(\cdot)$ denotes cosine similarity.

\textbf{Why this matters:} Unlike REMO~\cite{wu2025reflection} which retrieves based on \textit{input} similarity ($\text{sim}(x_t, x_i)$), TextBFGS retrieves based on \textit{gradient/error} similarity ($\text{sim}(g_t, g_i)$). For instance, an ``Index Error'' gradient in a sorting algorithm can retrieve a ``Boundary Check'' operator originally learned from a string parsing task, effectively transferring debugging logic across distinct coding domains.

\subsection{One-Pass Update Loop (Reuse and Revise)}

To ensure efficiency, TextBFGS fuses gradient generation and variable modification into a single LLM inference step, explicitly mirroring the \textbf{Reuse} and \textbf{Revise} phases of CBR. We design a specialized prompt structure forcing the LLM to explicitize the error and operator.

Let $\mathcal{P}$ be the system prompt. The update rule is formally:

\begin{equation}
g_t, \mathcal{O}_{t}, x_{t+1} = \text{LLM}(\mathcal{P}(x_t, E_t, \mathcal{O}_{\text{ref}}))
\end{equation}

Where $E_t$ is the execution result. The model performs the following internal Chain of Thought (CoT)~\cite{wei2022chain}:

\begin{enumerate}
    \item \textbf{Gradient Analysis ($g_t$):} The model analyzes the error pattern between the current output and the expectation (identifying the target problem).
    \item \textbf{Operator Application ($\mathcal{O}_{t}$):} The model generates the hybrid operator from $g_t$ and the retrieved set $\mathcal{O}_{\text{ref}}$, \textbf{Reusing} the historical case solutions.
    \item \textbf{Newton Update ($x_{t+1}$):} The model applies this operator to $x_t$, \textbf{Revising} the code to resolve the error.
\end{enumerate}

This mechanism reduces computational cost by 50\% compared to standard methods~\cite{xu2025metatextgrad} while injecting experience-driven second-order guidance.

\subsection{Case Retention: Self-Evolving Case Base}

TextBFGS mimics the self-correcting nature of Quasi-Newton methods by dynamically updating its HPCB during optimization, which corresponds to the \textbf{Retain} phase in CBR. In numerical optimization, BFGS achieves self-correction by iteratively refining its Hessian approximation using historical changes. TextBFGS transfers this principle: its HPCB continuously self-improves through accumulating error-operator pairs, making the curvature estimation increasingly accurate.

Upon a successful update (e.g., $f(x_{t+1}) > f(x_t)$), the system abstracts the specific transformation into a generalized operator $\mathcal{O}_{t}$. The resulting case tuple $(g_t, \mathcal{O}_{t})$ is then retained into the HPCB. This continuous retention enriches the case coverage, transforming the optimizer into a self-evolving agent that progressively masters the debugging landscape~\cite{madaan2022memory}.

\section{Experiments}

We empirically evaluate TextBFGS on the task of code optimization, where code serves as the executable text, demanding rigorous logic and structural correctness. We focus on cases where the base model fails initially, to strictly measure the optimizer's ability to turn failure into success through experience reuse.

\subsection{Experimental Setup}

\paragraph{Base Model \& Initialization.}

We employ Qwen3-235B-A22B~\cite{yang2025qwen3} as the backbone model for the optimizer, Qwen3-Embedding-8B~\cite{zhang2025qwen3} as the retrieval model, and ChromaDB\footnote{\url{https://github.com/chroma-core/chroma}} as the Case Base (CB), with the number of iterations limited to 20 per optimization task. To simulate a challenging debugging landscape and eliminate interference from internal reasoning traces, we disable the reasoning mode by setting enable-thinking to False, forcing the model to rely solely on explicit error feedback (gradients) rather than its internal latent reasoning. Additionally, we set the backbone model's temperature to 0.7, top-p to 0.95, and an appropriate max-tokens to ensure responses are not truncated. Other experimental details such as prompt implementation can be found in our \textit{code repository}\footnote{\url{https://github.com/TzuchengChang/TextBFGS}}.

\paragraph{Benchmarks \& Metrics.}

We employ EvalPlus\footnote{\url{https://github.com/evalplus/evalplus}}~\cite{evalplus} as the evaluation toolkit, which incorporates two standard benchmarks: HumanEval~\cite{chen2021evaluating} and MBPP~\cite{austin2021program}. To ensure the robustness of our results, we report the following metrics: 
\begin{itemize} 
    \item \textbf{Base Pass Rate:} The average pass rate of tasks on the standard test cases of HumanEval or MBPP. 
    \item \textbf{Plus Pass Rate:} The average pass rate on the augmented test cases provided by EvalPlus. These represent a more rigorous set of tasks designed to detect overfitting and evaluate the model's capability in handling edge cases. 
\end{itemize}

We filtered the datasets to retain only the problems where Qwen3-235B-A22B scored 0 in the initial Pass@1 attempt for further optimization. The filtered subsets are as follows: 
\begin{itemize} 
    \item \textbf{HumanEval-Hard:} 45 tasks (filtered from 164 tasks). 
    \item \textbf{MBPP-Hard:} 117 tasks (filtered from 378 tasks). 
\end{itemize}

Unless otherwise specified, HumanEval or MBPP mentioned in subsequent experiments refer to the above hard subsets of HumanEval or MBPP, and a task is considered failed if it fails to pass even a single testcase.

\paragraph{Baselines.}

To rigorously evaluate the effectiveness of TextBFGS, we compare it against three categories of approaches:
\begin{itemize}
    \item \textbf{TextGrad} and \textbf{TextGrad-Momentum}\footnote{\url{https://github.com/zou-group/textgrad}}~\cite{yuksekgonul2024textgrad}: The mainstream stateless feedback-driven textual optimization frameworks. TextGrad performs memoryless updates. TextGrad-Momentum extends this by aggregating a buffer of historical feedback to smooth the update trajectory, though it remains unaware of the global semantic geometry and lacks a persistent Case Base.
    
    \item \textbf{TextBFGS (w/o CB)}: An ablation variant of our framework where the Hessian-proxy case retrieval module is disabled. This baseline utilizes the same efficient One-Pass inference architecture to generate errors and updates simultaneously but operates without access to the Case Base. 
    
    \item \textbf{TextBFGS-REMO:} A TextBFGS reproduction of REMO~\cite{wu2025reflection}, the latest memory-augmented textual optimization framework. Since the official implementation is unavailable, we replicate it using the TextBFGS backbone and One-Pass architecture. Crucially, it employs problem-based retrieval: it queries the Case Base using the semantic similarity of the target problem input, simulating standard RAG-enhanced generation.
\end{itemize}

\paragraph{Case Base Construction.}

To populate the Hessian-proxy Case Base with high-quality data, we employ a bootstrapping procedure using the stateless variant, TextBFGS (w/o CB). We execute it on both HumanEval-Hard and MBPP-Hard with 3 optimization learning epochs per task.

For each successful optimization step, we store a case tuple containing 4 components: the pre-optimization text ($x_t$), the error description/gradient ($g_t$), the abstract adaptation rule/operator ($\mathcal{O}_t$), and the post-optimization text ($x_{t+1}$). This unified schema supports both our method (which queries via $g_t$) and the TextBFGS-REMO baseline (which queries via the surface-level problem $x_t$).

This process yields a concentrated repository of 39 case trajectories for the HumanEval domain and 245 for the MBPP domain. During inference, we retrieve the top-$k$ ($k=3$) most similar entries. The choice of $k=3$ aligns exactly with the TextGrad-Momentum baseline, which maintains a context window of the three most recent optimization records, ensuring a fair comparison of context length and information capacity.

\subsection{Ablation Study: Impact of CBR Components}

To dissect the contribution of each module in TextBFGS, we conduct a component-wise analysis on both HumanEval-Hard and MBPP-Hard. For consistency in this ablation, both tasks utilize the Case Base (CB) from MBPP, which contains richer cases than HumanEval. The results are presented in Table \ref{tab:ablation_humaneval} and Table \ref{tab:ablation_mbpp}.

\begin{table}[ht]
\centering
\begin{tabular}{l | cc}
\toprule
\textbf{Method} & \textbf{Base Pass} & \textbf{Plus Pass} \\
\midrule
TextGrad & 91.11\% & 82.22\% \\
TextGrad-Momentum & 91.11\% & 86.67\% \\
\midrule
TextBFGS (w/o CB) & 91.11\% & 82.22\% \\
TextBFGS-REMO & 95.56\% & 91.11\% \\
\textbf{TextBFGS} & \textbf{97.78\%} & \textbf{93.33\%} \\
\bottomrule
\end{tabular}
\caption{Ablation on HumanEval-Hard (using MBPP CB). Case Base integration yields substantial gains. Notably, retrieving based on error gradients (TextBFGS) outperforms problem-based retrieval (TextBFGS-REMO) by +2.22\% in both Pass Rate}
\label{tab:ablation_humaneval}
\end{table}

\begin{table}[ht]
\centering
\begin{tabular}{l | cc}
\toprule
\textbf{Method} & \textbf{Base Pass} & \textbf{Plus Pass} \\
\midrule
TextGrad & 85.47\% & 48.72\% \\
TextGrad-Momentum & 88.89\% & 58.12\% \\
\midrule
TextBFGS (w/o CB) & 85.47\% & 48.72\% \\
TextBFGS-REMO & \textbf{95.73\%} & \textbf{78.63\%} \\
\textbf{TextBFGS} & 94.02\% & 74.36\% \\
\bottomrule
\end{tabular}
\caption{Ablation on MBPP-Hard (using MBPP CB). When the case source strictly aligns with the target task, problem-based retrieval (TextBFGS-REMO) performs slightly better, though error-based retrieval (TextBFGS) remains highly competitive.}
\label{tab:ablation_mbpp}
\end{table}

\paragraph{Impact of Momentum.}

Comparing standard TextGrad with TextGrad-Momentum, we observe that aggregating historical feedback consistently improves robustness. Especially on MBPP-Hard (Table \ref{tab:ablation_mbpp}), momentum yields a boost of +9.4\% in Plus pass rate (from 48.72\% to 58.12\%). This confirms that recording the update direction helps prevent the optimizer from oscillating in complex solution spaces.

\paragraph{Impact of Case Base Augmentation.}

The most significant performance leap stems from the integration of the Hessian-Proxy Case Base. Regardless of the retrieval strategy, retrieval methods drastically outperform stateless baselines. For instance, on HumanEval-Hard, the Plus pass rate improved by up to \textbf{+6.66\%} compared with Momentum (from 86.67\% to 93.33\%), while on MBPP-Hard, the Plus pass rate showed a maximum increase of \textbf{+20.51\%} compared with Momentum (from 58.12\% to 78.63\%). This validates our premise that accessing historical successful cases is essential for solving hard instances.

\subsection{Cross-Domain vs.\ In-Domain Retrieval}

To strictly evaluate the generalization capability of different retrieval methods, we compare performance when the Case Base (CB) is populated with data from the same domain (In-Domain) versus a disjoint domain (Cross-Domain). This experiment tests whether the system relies on memorizing task-specific code snippets or learning transferable debugging logic. We split the results by retrieval strategy into Table~\ref{tab:cross_domain_remo} and Table~\ref{tab:cross_domain_gradient} for clarity.

\begin{table}[ht]
\centering
\begin{tabular}{l | cc}
\toprule
\textbf{Target Task \& Case Source} & \textbf{Base Pass} & \textbf{Plus Pass} \\
\midrule
HumanEval \ (w/o CB) & 91.11\% & 82.22\% \\
MBPP \ (w/o CB) & 85.47\% & 48.72\% \\
\midrule
HumanEval \ (HumanEval CB, In) & \textbf{99.39\%} & \textbf{97.56\%} \\
HumanEval \ (MBPP CB, Cross) & 95.56\% & 91.11\% \\
MBPP \ (MBPP CB, In) & \textbf{95.73\%} & \textbf{78.63\%} \\
MBPP \ (HumanEval CB, Cross) & 91.45\% & 58.12\% \\
\bottomrule
\end{tabular}
\caption{Performance of TextBFGS-REMO (problem-based retrieval) degrades significantly in Cross-Domain scenarios. Notably, on MBPP using the HumanEval CB, the robust Plus metric drops -20.51\%, indicating poor transferability.}
\label{tab:cross_domain_remo}
\end{table}

\begin{table}[ht]
\centering
\begin{tabular}{l | cc}
\toprule
\textbf{Target Task \& Case Source} & \textbf{Base Pass} & \textbf{Plus Pass} \\
\midrule
HumanEval \ (w/o CB) & 91.11\% & 82.22\% \\
MBPP \ (w/o CB) & 85.47\% & 48.72\% \\
\midrule
HumanEval \ (HumanEval CB, In) & \textbf{99.39\%} & \textbf{96.34\%} \\
HumanEval \ (MBPP CB, Cross) & 97.78\% & 93.33\% \\
MBPP \ (MBPP CB, In) & 94.02\% & 74.36\% \\
MBPP \ (HumanEval CB, Cross) & \textbf{94.02\%} & \textbf{74.36\%} \\
\bottomrule
\end{tabular}
\caption{Performance of TextBFGS (error-based retrieval) maintains high accuracy across domains. Crucially, when optimizing MBPP using HumanEval CB, TextBFGS significantly outperforms TextBFGS-REMO by +16.24\% on the Plus.}
\label{tab:cross_domain_gradient}
\end{table}

\paragraph{Analysis.} 

As shown in Table~\ref{tab:cross_domain_remo}, the problem-based retrieval (TextBFGS-REMO) exhibits signs of surface-level case overfitting. While it performs slightly better in In-Domain settings (e.g., 97.56\% vs.\ 96.34\% on HumanEval) by retrieving near-identical historical examples, it fails to transfer knowledge across disjoint tasks. Specifically, when solving MBPP tasks using HumanEval cases, the Plus pass rate collapses to 58.12\%, barely improving over the baseline (48.72\%).

In contrast, Table~\ref{tab:cross_domain_gradient} demonstrates the superior generalization of TextBFGS. By retrieving cases based on error dynamics (gradients) rather than surface problem descriptions, TextBFGS successfully identifies shared failure modes even when the task contexts differ. Consequently, on the challenging MBPP benchmark with HumanEval cases, TextBFGS maintains a high Plus pass rate of 74.36\%, outperforming the input-based baseline by a substantial margin of \textbf{+16.24\%}. This confirms that optimization logic, encapsulated as adaptation operators in the case base, is more transferable than specific code instances.

\subsection{Efficiency Analysis}
\label{sec:efficiency}

We assess the computational cost on both datasets and utilize the Case Base (CB) from MBPP to evaluate efficiency under cross-domain and in-domain settings. We employ an early-stopping mechanism. Consequently, a stronger solver that finds solutions faster will lead to faster convergence, naturally resulting in fewer average API calls. The results are presented in Table~\ref{tab:eff_he} and Table~\ref{tab:eff_mbpp}. We summarize the key observations below:

\begin{table}[ht]
\centering
\begin{tabular}{l c c c}
\toprule
\textbf{Method} & \textbf{Calls/Task} & \textbf{Tokens/Call} & \textbf{Tokens/Task}\\
\midrule
TextGrad & 35.8 & \textbf{863.9} & 30.9k\\
TextGrad-Momentum & 29.8 & 1464.2 & 43.7k\\
\midrule
TextBFGS (w/o CB) & 17.0 & 1481.3 & 25.2k\\
TextBFGS-REMO & 13.9 & 1594.4 & 22.2k\\
\textbf{TextBFGS} & \textbf{13.6} & 1581.7 & \textbf{21.6k}\\
\bottomrule
\end{tabular}
\caption{Efficiency on HumanEval-Hard (Cross-Domain). Using MBPP CB, TextBFGS achieves the best trade-off, demonstrating superior efficiency in transferring debugging logic across domains.}
\label{tab:eff_he}
\end{table}

\begin{table}[ht]
\centering
\begin{tabular}{l c c c}
\toprule
\textbf{Method} & \textbf{Calls/Task} & \textbf{Tokens/Call} & \textbf{Tokens/Task}\\
\midrule
TextGrad & 36.6 & \textbf{727.3} & 26.6k\\
TextGrad-Momentum & 34.0 & 1114.1 & 37.9k\\
\midrule
TextBFGS (w/o CB) & 18.2 & 1330.2 & 24.2k\\
TextBFGS-REMO & \textbf{7.1} & 2262.0 & \textbf{16.1k}\\
\textbf{TextBFGS} & 8.2 & 2103.9 & 17.2k\\
\bottomrule
\end{tabular}
\caption{Efficiency on MBPP-Hard (In-Domain). In-Domain retrieval results in higher token consumption per call compared to Table~\ref{tab:eff_he} due to the retrieval of more detailed, context-heavy specific experiences, but it sharply reduces the number of calls.}
\label{tab:eff_mbpp}
\end{table}

\paragraph{One-Pass Inference Reduces API Overhead.} 

A distinct advantage of our framework is the One-Pass architecture. Methods like TextGrad and REMO operate in a two-stage manner: one API call to compute the error diagnosis (gradient) and a second separate call to apply the update. In contrast, TextBFGS integrates the diagnosis computation (via internal CoT) and the operator execution into a single inference step.

As shown in Table~\ref{tab:eff_he}, although TextGrad has the lowest tokens per call (863.9), it requires significantly more API calls (35.8). While both TextBFGS (w/o CB) and TextGrad achieve identical performance (82.22\%), TextBFGS (w/o CB) achieves an \textbf{18.4\%} cost reduction compared to TextGrad (25.2k vs. 30.9k), proving that One-Pass inference saves substantial overhead.

\paragraph{Token Consumption and Retrieval Granularity.}

We analyze the trade-off using Tokens/Task. As shown in Table~\ref{tab:eff_he} and Table~\ref{tab:eff_mbpp}, two distinct patterns emerge:
\begin{itemize}
    \item \textbf{Inefficiency of Stateless Momentum:} While TextGrad-Momentum improves pass rates over vanilla TextGrad, it incurs the highest computational cost (43.7k). This is because Momentum simply concatenates the optimization history. Since this history is stateless and unfiltered, it may contain irrelevant or redundant information that bloats the context without proportionally reducing the number of iterations.
    
    \item \textbf{Efficiency via Case-Retrieval:} In contrast, although retrieving cases increases the input size, the injected operators are highly relevant "high-utility" adaptation priors that guide the model to the correct solution much faster. This drastically cuts the iteration count (e.g. down to 13.6), achieving higher performance (from 86.67\% to 93.33\%), resulting in the lowest total token consumption (21.6k), effectively saving a total of \textbf{50.6\%} tokens compared to Momentum.
\end{itemize}

Furthermore, we observe a granularity preference across domains. Case-retrieval methods consume obviously fewer tokens per call on HumanEval (Cross-Domain, 1582 tokens) than on MBPP (In-Domain, 2104 tokens). We attribute this to the nature of case matching: In-domain queries tend to fetch lengthy, task-specific snippets, whereas cross-domain queries retrieve shorter, more abstract structural logic. This suggests that in cross-domain scenarios, generalizable debugging experiences matter most, and our system naturally adapts its retrieval granularity, maintaining efficiency even when transferring knowledge across disjoint domains.

\section{Case Study}

We present a concrete example from our real optimization logs: \texttt{HumanEval/127}. This case directly illustrates that TextBFGS can turn a persistent failure mode into a fully correct solution under the same backbone model.

\paragraph{Initial code (before optimization).}
\begin{small}
\begin{verbatim}
print(intersection((1, 2), (2, 3)))       # Output: "NO"
print(intersection((-1, 1), (0, 4)))      # Output: "NO"
print(intersection((-3, -1), (-5, 5)))    # Output: "YES"
\end{verbatim}
\end{small}

\paragraph{TextGrad result (still fails).}
\begin{small}
\begin{verbatim}
length = intersect_end - intersect_start + 1  # Inclusive interval
...
return "YES" if is_prime(length) else "NO"
\end{verbatim}
\end{small}

\paragraph{TextBFGS result (passes all tests).}
\begin{small}
\begin{verbatim}
length = overlap_end - overlap_start  # corrected difference
...
return "YES" if is_prime(length) else "NO"
\end{verbatim}
\end{small}

Compared with TextGrad, TextBFGS retrieves and applies a more precise correction pattern for this boundary-sensitive logic. As a result, it fixes the off-by-one error and achieves full pass on both base and plus test suites.

\section{Conclusion and Future Work}

We introduced TextBFGS, a Case-Based Reasoning (CBR) framework inspired by Quasi-Newton methods to overcome the stateless inefficiencies of first-order code optimization. By retrieving historical error-operators instead of surface-level problems, TextBFGS conceptually approximates semantic curvature, enabling the robust cross-domain transfer of debugging logic. Coupled with efficient One-Pass updates and Case Retention, it establishes a highly effective, experience-driven paradigm for LLM self-correction. Future work will explore extending this framework beyond code generation to broader reasoning tasks and dynamic agentic workflows.

\newpage
\begin{credits}
\subsubsection{Declaration on Generative AI}
We utilized Large Language Models (e.g., DeepSeek~\cite{liu2024deepseek}, Gemini~\cite{team2023gemini}) solely for grammatical refinement and polishing of the text. All scientific claims and experimental results are the authors' own work.
\end{credits}

\bibliographystyle{splncs04}
\bibliography{bibliography}

@inproceedings{yang2023large,
  title={Large language models as optimizers},
  author={Yang, Chengrun and Wang, Xuezhi and Lu, Yifeng and Liu, Hanxiao and Le, Quoc V and Zhou, Denny and Chen, Xinyun},
  booktitle={The Twelfth International Conference on Learning Representations},
  year={2023}
}

@article{opsahl2024optimizing,
  title={Optimizing instructions and demonstrations for multi-stage language model programs},
  author={Opsahl-Ong, Krista and Ryan, Michael J and Purtell, Josh and Broman, David and Potts, Christopher and Zaharia, Matei and Khattab, Omar},
  journal={arXiv preprint arXiv:2406.11695},
  year={2024}
}

@article{agrawal2025gepa,
  title={Gepa: Reflective prompt evolution can outperform reinforcement learning},
  author={Agrawal, Lakshya A and Tan, Shangyin and Soylu, Dilara and Ziems, Noah and Khare, Rishi and Opsahl-Ong, Krista and Singhvi, Arnav and Shandilya, Herumb and Ryan, Michael J and Jiang, Meng and others},
  journal={arXiv preprint arXiv:2507.19457},
  year={2025}
}

@article{amari1993backpropagation,
  title={Backpropagation and stochastic gradient descent method},
  author={Amari, Shun-ichi},
  journal={Neurocomputing},
  volume={5},
  number={4-5},
  pages={185--196},
  year={1993},
  publisher={Elsevier}
}

@article{liu1989limited,
  title={On the limited memory BFGS method for large scale optimization},
  author={Liu, Dong C and Nocedal, Jorge},
  journal={Mathematical programming},
  volume={45},
  number={1},
  pages={503--528},
  year={1989},
  publisher={Springer}
}

@inproceedings{khattab2024dspy,
  title={DSPy: Compiling Declarative Language Model Calls into Self-Improving Pipelines},
  author={Khattab, Omar and Singhvi, Arnav and Maheshwari, Paridhi and Zhang, Zhiyuan and Santhanam, Keshav and Vardhamanan, Sri and Haq, Saiful and Sharma, Ashutosh and Joshi, Thomas T. and Moazam, Hanna and Miller, Heather and Zaharia, Matei and Potts, Christopher},
  journal={The Twelfth International Conference on Learning Representations},
  year={2024}
}

@article{fernando2023promptbreeder,
  title={Promptbreeder: Self-referential self-improvement via prompt evolution},
  author={Fernando, Chrisantha and Banarse, Dylan and Michalewski, Henryk and Osindero, Simon and Rockt{\"a}schel, Tim},
  journal={arXiv preprint arXiv:2309.16797},
  year={2023}
}

@article{guo2023connecting,
  title={Connecting large language models with evolutionary algorithms yields powerful prompt optimizers},
  author={Guo, Qingyan and Wang, Rui and Guo, Junliang and Li, Bei and Song, Kaitao and Tan, Xu and Liu, Guoqing and Bian, Jiang and Yang, Yujiu},
  journal={arXiv preprint arXiv:2309.08532},
  year={2023}
}

@article{novikov2025alphaevolve,
  title={AlphaEvolve: A coding agent for scientific and algorithmic discovery},
  author={Novikov, Alexander and V{\~u}, Ng{\^a}n and Eisenberger, Marvin and Dupont, Emilien and Huang, Po-Sen and Wagner, Adam Zsolt and Shirobokov, Sergey and Kozlovskii, Borislav and Ruiz, Francisco JR and Mehrabian, Abbas and others},
  journal={arXiv preprint arXiv:2506.13131},
  year={2025}
}

@article{madaan2023self,
  title={Self-refine: Iterative refinement with self-feedback},
  author={Madaan, Aman and Tandon, Niket and Gupta, Prakhar and Hallinan, Skyler and Gao, Luyu and Wiegreffe, Sarah and Alon, Uri and Dziri, Nouha and Prabhumoye, Shrimai and Yang, Yiming and others},
  journal={Advances in Neural Information Processing Systems},
  volume={36},
  pages={46534--46594},
  year={2023}
}

@article{shinn2023reflexion,
  title={Reflexion: Language agents with verbal reinforcement learning},
  author={Shinn, Noah and Cassano, Federico and Gopinath, Ashwin and Narasimhan, Karthik and Yao, Shunyu},
  journal={Advances in Neural Information Processing Systems},
  volume={36},
  pages={8634--8652},
  year={2023}
}

@article{zhang2024revolve,
  title={Revolve: Optimizing ai systems by tracking response evolution in textual optimization},
  author={Zhang, Peiyan and Jin, Haibo and Hu, Leyang and Li, Xinnuo and Kang, Liying and Luo, Man and Song, Yangqiu and Wang, Haohan},
  journal={arXiv preprint arXiv:2412.03092},
  year={2024}
}

@article{zhanghessiangrad,
  title={HessianGrad: Optimizing AI Systems with Hessian-Aware Textual Gradients},
  author={Zhang, Peiyan and Jin, Haibo and Hu, Leyang and Li, Xinnuo and Kang, Liying and Luo, Man and Song, Yangqiu and Wang, Haohan},
  year={2024}
}

@article{suzgun2024meta,
  title={Meta-prompting: Enhancing language models with task-agnostic scaffolding},
  author={Suzgun, Mirac and Kalai, Adam Tauman},
  journal={arXiv preprint arXiv:2401.12954},
  year={2024}
}

@article{lewis2020retrieval,
  title={Retrieval-augmented generation for knowledge-intensive nlp tasks},
  author={Lewis, Patrick and Perez, Ethan and Piktus, Aleksandra and Petroni, Fabio and Karpukhin, Vladimir and Goyal, Naman and K{\"u}ttler, Heinrich and Lewis, Mike and Yih, Wen-tau and Rockt{\"a}schel, Tim and others},
  journal={Advances in neural information processing systems},
  volume={33},
  pages={9459--9474},
  year={2020}
}

@article{ethayarajh2019contextual,
  title={How contextual are contextualized word representations},
  author={Ethayarajh, Kawin},
  journal={Comparing the geometry of BERT, ELMo, and GPT-2 Embeddings},
  volume={2},
  year={2019}
}

@inproceedings{gao2025devil,
  title={The devil is in the prompts: Retrieval-augmented prompt optimization for text-to-video generation},
  author={Gao, Bingjie and Gao, Xinyu and Wu, Xiaoxue and Zhou, Yujie and Qiao, Yu and Niu, Li and Chen, Xinyuan and Wang, Yaohui},
  booktitle={Proceedings of the Computer Vision and Pattern Recognition Conference},
  pages={3173--3183},
  year={2025}
}

@article{zhang2016introduction,
  title={Introduction to machine learning: k-nearest neighbors},
  author={Zhang, Zhongheng},
  journal={Annals of translational medicine},
  volume={4},
  number={11},
  pages={218},
  year={2016}
}

@article{wei2022chain,
  title={Chain-of-thought prompting elicits reasoning in large language models},
  author={Wei, Jason and Wang, Xuezhi and Schuurmans, Dale and Bosma, Maarten and Xia, Fei and Chi, Ed and Le, Quoc V and Zhou, Denny and others},
  journal={Advances in neural information processing systems},
  volume={35},
  pages={24824--24837},
  year={2022}
}

@article{xu2025metatextgrad,
  title={metaTextGrad: Automatically optimizing language model optimizers},
  author={Xu, Guowei and Yuksekgonul, Mert and Guestrin, Carlos and Zou, James},
  journal={arXiv preprint arXiv:2505.18524},
  year={2025}
}

@article{madaan2022memory,
  title={Memory-assisted prompt editing to improve GPT-3 after deployment},
  author={Madaan, Aman and Tandon, Niket and Clark, Peter and Yang, Yiming},
  journal={arXiv preprint arXiv:2201.06009},
  year={2022}
}

@article{yang2025qwen3,
  title={Qwen3 technical report},
  author={Yang, An and Li, Anfeng and Yang, Baosong and Zhang, Beichen and Hui, Binyuan and Zheng, Bo and Yu, Bowen and Gao, Chang and Huang, Chengen and Lv, Chenxu and others},
  journal={arXiv preprint arXiv:2505.09388},
  year={2025}
}

@article{zhang2025qwen3,
  title={Qwen3 Embedding: Advancing Text Embedding and Reranking Through Foundation Models},
  author={Zhang, Yanzhao and Li, Mingxin and Long, Dingkun and Zhang, Xin and Lin, Huan and Yang, Baosong and Xie, Pengjun and Yang, An and Liu, Dayiheng and Lin, Junyang and others},
  journal={arXiv preprint arXiv:2506.05176},
  year={2025}
}

@inproceedings{evalplus,
  title = {Is Your Code Generated by Chat{GPT} Really Correct? Rigorous Evaluation of Large Language Models for Code Generation},
  author = {Liu, Jiawei and Xia, Chunqiu Steven and Wang, Yuyao and Zhang, Lingming},
  booktitle = {Thirty-seventh Conference on Neural Information Processing Systems},
  year = {2023},
  url = {https://openreview.net/forum?id=1qvx610Cu7},
}

@article{chen2021evaluating,
  title={Evaluating large language models trained on code},
  author={Chen, Mark},
  journal={arXiv preprint arXiv:2107.03374},
  year={2021}
}

@article{austin2021program,
  title={Program synthesis with large language models},
  author={Austin, Jacob and Odena, Augustus and Nye, Maxwell and Bosma, Maarten and Michalewski, Henryk and Dohan, David and Jiang, Ellen and Cai, Carrie and Terry, Michael and Le, Quoc and others},
  journal={arXiv preprint arXiv:2108.07732},
  year={2021}
}

@article{yuksekgonul2024textgrad,
  title={Textgrad: Automatic" differentiation" via text},
  author={Yuksekgonul, Mert and Bianchi, Federico and Boen, Joseph and Liu, Sheng and Huang, Zhi and Guestrin, Carlos and Zou, James},
  journal={arXiv preprint arXiv:2406.07496},
  year={2024}
}

@article{wu2025reflection,
  title={Reflection-Enhanced Meta-Optimization Integrating TextGrad-style Prompt Optimization with Memory-Driven Self-Evolution},
  author={Wu, Chunlong and Qu, Zhibo},
  journal={arXiv preprint arXiv:2508.18749},
  year={2025}
}

@article{sahoo2024systematic,
  title={A systematic survey of prompt engineering in large language models: Techniques and applications},
  author={Sahoo, Pranab and Singh, Ayush Kumar and Saha, Sriparna and Jain, Vinija and Mondal, Samrat and Chadha, Aman},
  journal={arXiv preprint arXiv:2402.07927},
  year={2024}
}

@article{ma2025agentic,
  title={Agentic Neural Networks: Self-Evolving Multi-Agent Systems via Textual Backpropagation},
  author={Ma, Xiaowen and Lin, Chenyang and Zhang, Yao and Tresp, Volker and Ma, Yunpu},
  journal={arXiv preprint arXiv:2506.09046},
  year={2025}
}

@article{liu2024deepseek,
  title={Deepseek-v3 technical report},
  author={Liu, Aixin and Feng, Bei and Xue, Bing and Wang, Bingxuan and Wu, Bochao and Lu, Chengda and Zhao, Chenggang and Deng, Chengqi and Zhang, Chenyu and Ruan, Chong and others},
  journal={arXiv preprint arXiv:2412.19437},
  year={2024}
}

@article{team2023gemini,
  title={Gemini: a family of highly capable multimodal models},
  author={Team, Gemini and Anil, Rohan and Borgeaud, Sebastian and Alayrac, Jean-Baptiste and Yu, Jiahui and Soricut, Radu and Schalkwyk, Johan and Dai, Andrew M and Hauth, Anja and Millican, Katie and others},
  journal={arXiv preprint arXiv:2312.11805},
  year={2023}
}

\end{document}